\documentclass[10pt, conference, compsocconf]{IEEEtran}
\ifCLASSINFOpdf
\else
\fi
\usepackage{amsmath}
\usepackage{textcomp}
\usepackage{gensymb}
\usepackage{hyperref}
\usepackage{multirow}
\usepackage{booktabs}
\usepackage{tabularx}
\usepackage{float}
\usepackage{booktabs}
\usepackage{svg}
\usepackage{subcaption}
\usepackage{graphicx}
\usepackage{pifont}
\usepackage{svg}
\usepackage{algorithm}
\usepackage{algpseudocode}
\usepackage{tikz}
\hypersetup{
    colorlinks=true,
    linkcolor=blue,
    filecolor=magenta,      
    urlcolor=cyan,
    pdftitle={D2AHMR},
    pdfpagemode=FullScreen,
    }
\begin{document}
%
\title{Distribution and Depth-Aware Transformers for 3D Human Mesh Recovery}

\author{\IEEEauthorblockN{Jerrin Bright\IEEEauthorrefmark{1},
Bavesh Balaji\IEEEauthorrefmark{1},
Harish Prakash\IEEEauthorrefmark{1},
Yuhao Chen\IEEEauthorrefmark{1},
David A Clausi\IEEEauthorrefmark{1} and
John Zelek\IEEEauthorrefmark{1}}
\IEEEauthorblockA{\IEEEauthorrefmark{1} Vision and Image Processing Lab, University of Waterloo, Canada}
{\{jerrin.bright, bbalaji, harish.prakash, yuhao.chen1, dclausi, jzelek\}}@uwaterloo.ca

}

\maketitle

\begin{abstract}
   Precise Human Mesh Recovery (HMR) with in-the-wild data is a formidable challenge and is often hindered by depth ambiguities and reduced precision. Existing works resort to either pose priors or multi-modal data such as multi-view or point cloud information, though their methods often overlook the valuable scene-depth information inherently present in a single image. Moreover, achieving robust HMR for out-of-distribution (OOD) data is exceedingly challenging due to inherent variations in pose, shape and depth. Consequently, understanding the underlying distribution becomes a vital subproblem in modeling human forms. Motivated by the need for unambiguous and robust human modeling, we introduce Distribution and depth-aware human mesh recovery (D2A-HMR), an end-to-end transformer architecture meticulously designed to minimize the disparity between distributions and incorporate scene-depth leveraging prior depth information. Our approach demonstrates superior performance in handling OOD data in certain scenarios while consistently achieving competitive results against state-of-the-art HMR methods on controlled datasets. 
\end{abstract}


%
\IEEEpeerreviewmaketitle

\section{Introduction}\label{sec:intro}

Monocular Human Mesh Recovery (HMR) is an approach for estimating the pose and shape of a human subject from a single image, featuring a broad spectrum of applications in various downstream tasks \cite{econ, neuralbody, mime}. HMR can be split into two types: parametric and non-parametric approaches. The parametric approach involves the modeling of a network to generate model parameters, which are subsequently utilized for human mesh generation, as elucidated in \cite{hmr, hmmr, imphmr}. Recent strides have been witnessed in nonparametric-based approaches \cite{pose2mesh, metro}, which directly regresses the 3D coordinates of the human mesh.

Despite the notable progress in both paradigms, they struggle with two key challenges- the appearance domain gap and depth ambiguity. Controlled environments, often used for training, offer a setting where data collection and annotation are manageable and precise. However, the challenge arises when the trained model is applied to in-the-wild data, where real-world variability, such as lighting conditions, backgrounds, and poses, differs significantly from controlled settings. Second, depth-ambiguity issues plague single-view images. In response to the latter challenge, researchers, as exemplified in \cite{vibe} and \cite{hmmr}, have proposed solutions that leverage temporal information extracted from video inputs to enhance the understanding of human motion. However, these temporal approaches have entailed significant computational overhead.

\begin{figure}[t]
  \centering
   \includegraphics[width=\linewidth]{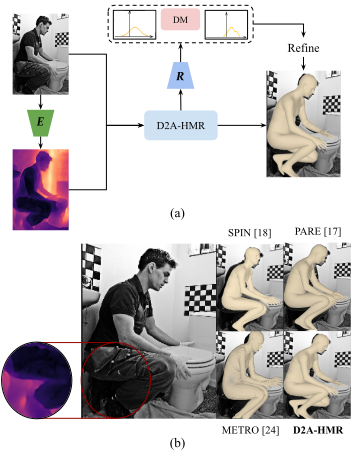}
   \vspace{-2em}
   \caption{\textbf{Illustration of our main idea}. (a) Overview of the proposed D2A-HMR approach (b) Our method, D2A-HMR improves the mesh-image alignment (particularly as visualized in the highlighted region) when compared against SPIN \cite{spin}, PARE \cite{pare} and METRO \cite{metro}.} 
   \label{fig:teaser}
\end{figure}

Obtaining ground truth mesh labels for human mesh reconstruction is a tedious task, mainly due to challenges like complexities of dynamic human motion, scene dynamics, resource constraints, and privacy concerns. In response to the inherent difficulty in obtaining accurate ground truth labels, existing works such as \cite{metro, pose2mesh, spin} resort to using pseudo ground truth to train models. Consequently, the modeling of human forms is inherently biased due to the presence of noisy labels. Moreover, the generalization of HMR for OOD poses, as discussed earlier, is an immensely challenging problem. Prior works \cite{prohmr, nf_smpl1} model the output as a distribution of plausible 3D pose using normalizing flows and use information such as 2D keypoints or part segments as priors to provide deterministic predictions for downstream tasks. However, since these models use normalizing flows to explicitly estimate the underlying output distribution, they fail to generalize as shown in \cite{Kirichenko2020WhyNF} and do not solve the model's bias to actual data, especially in scenarios with noisy labels and uncertainties.  

To address the limitations of existing methods, our work introduces a novel approach to address these issues through a depth- and distribution-aware framework designed for the recovery of human mesh from monocular images. Notably, we integrate scene-depth information from monocular cameras obtained from previous depth models (termed \textit{pseudo-depth}) into a transformer encoder via the cross-attention mechanism. In addition, we employ a log-likelihood residual approach to learn deviations in the underlying distribution, facilitating a refinement module in the training process. This distribution approach \textit{explicitly encourages the model} to learn a more generalizable representation that can perform better on unseen data. To further refine the mesh shape and feature relationships, we introduce a dedicated silhouette decoder and a masked modeling module. As showcased in Figure \ref{fig:teaser}, these contributions allow our D2A-HMR approach to excel in \textit{handling challenging, unseen poses}. To the best of our knowledge, D2A-HMR is the only framework to explicitly incorporate depth priors and systematically learn the mesh distribution disparity between the underlying prediction and ground truth distributions. Through experimentation, we demonstrate that our method outperforms existing works on some benchmarked datasets. In summary, our contributions include:

\begin{enumerate}
    \item We introduce a novel image-based HMR model named {D2A-HMR} that adeptly models the underlying distributions and integrates pseudo-depth priors for efficient and accurate mesh recovery.

    \item By leveraging {residual log-likehood} approach, we refine the model by learning the disparity between the underlying predicted and ground truth distribution.
    
    \item Validation of the {enhanced performance} through the integration of pseudo-depth and distribution-aware modules in HMR, particularly in complex human pose scenarios.
\end{enumerate}

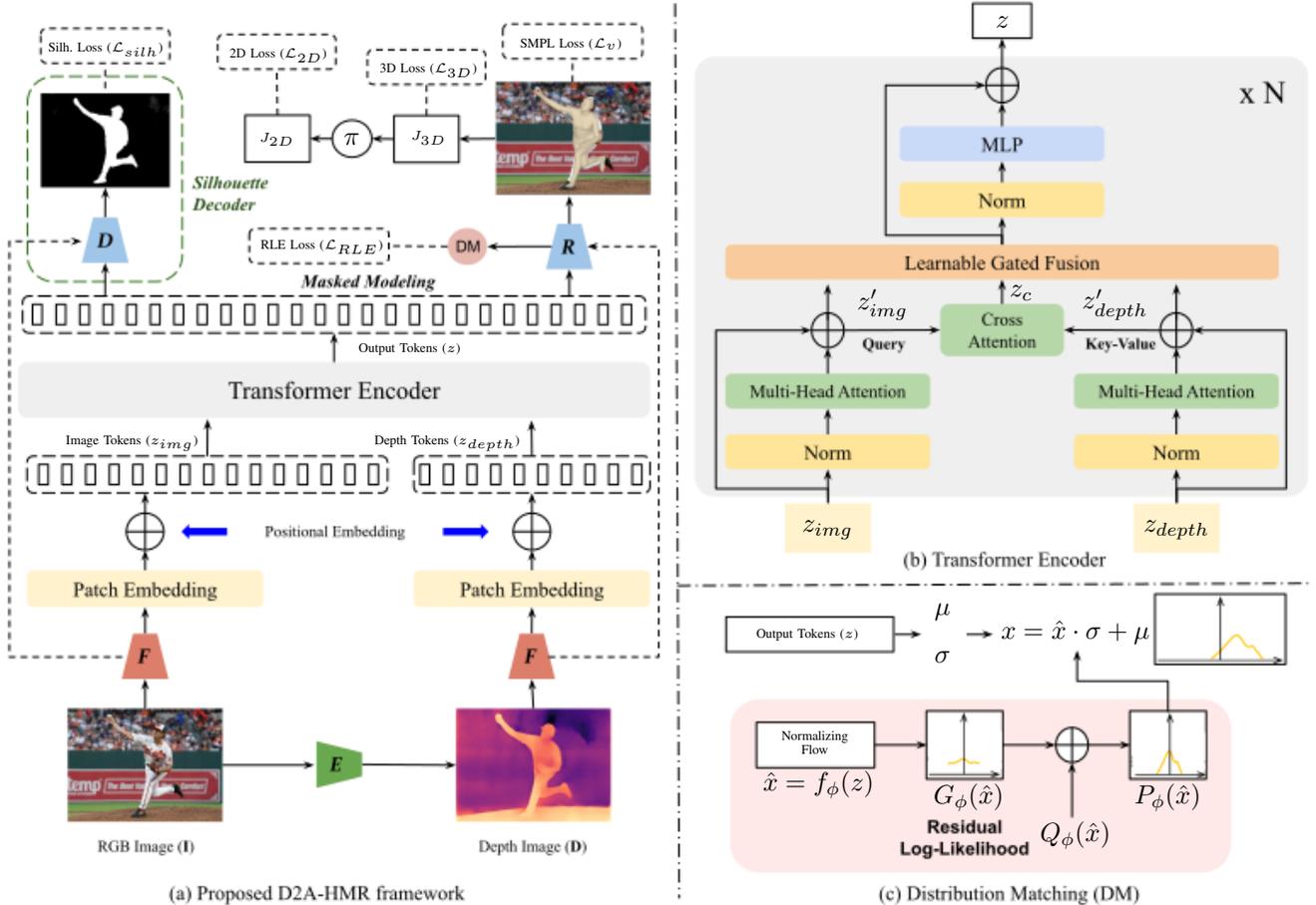
\begin{figure*}[t]
  \centering

  \begin{tikzpicture}
    \node at (0,0) {\includegraphics[width=\linewidth]{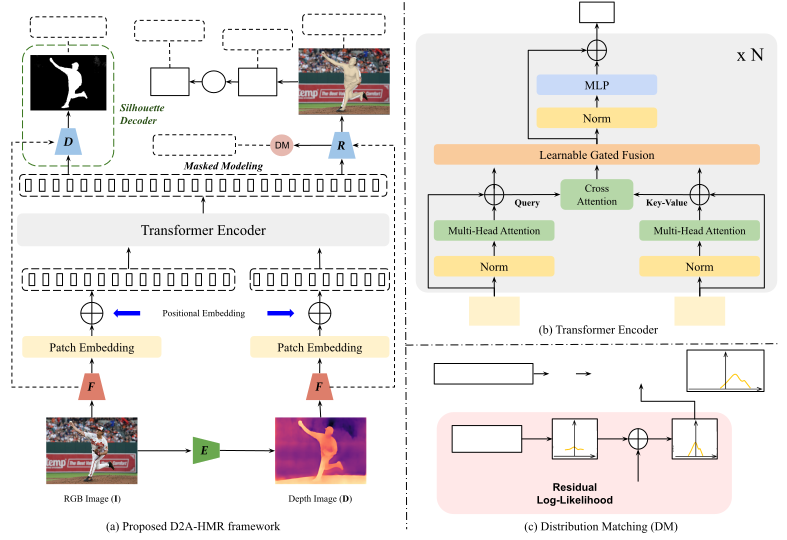}};
    \node[font=\tiny] at (-7.35,5.45) {Silh. Loss ($\mathcal{L}_{silh}$)}; 
    \node[font=\tiny] at (-5.05,5.37) {2D Loss ($\mathcal{L}_{2D}$)};
    \node[font=\tiny] at (-3.05,5.17) {3D Loss ($\mathcal{L}_{3D}$)};
    \node[font=\tiny] at (-1.15,5.525) {SMPL Loss ($\mathcal{L}_{v}$)};
    \node[font=\tiny] at (-3.05,4.27) {$J_{3D}$};
    \node at (-4.07,4.25) {$\pi$};
    \node[font=\tiny] at (-5.07,4.25) {$J_{2D}$};
    \node[font=\tiny] at (-4.47,2.82) {RLE Loss ($\mathcal{L}_{RLE}$)};
    \node[font=\tiny] at (-3.3,1.45) {Output Tokens ($z$)};
    \node[font=\tiny] at (-7,0.2) {Image Tokens ($z_{img}$)};
    \node[font=\tiny] at (-2.8,0.2) {Depth Tokens ($z_{depth}$)};
    
    \node at (4.6,5.81) {$z$};
    \node at (3,2.05) {$z'_{img}$};
    \node at (6.1,2.05) {$z'_{depth}$};
    \node at (4.85,2.2) {$z_{c}$};
    \node at (2.27,-0.95) {$z_{img}$};
    \node at (6.92,-0.95) {$z_{depth}$};
    \node at (3.8,-2.05) {$\mu$};
    \node at (3.8,-2.65) {$\sigma$};
    \node at (5.57,-2.325) {$x = \hat{x} \cdot \sigma + \mu $};
    
    \node[font=\tiny] at (2,-2.35) {Output Tokens ($z$)};
    \node[font=\tiny] at (2.1,-3.72) {Normalizing};
    \node[font=\tiny] at (2.1,-3.89) {Flow};
    \node at (2.15,-4.35) {$\hat{x} = f_\phi(z)$};
    \node at (4.15,-4.525) {$G_\phi(\hat{x})$};
    \node at (5.57,-5.025) {$Q_\phi(\hat{x})$};
    \node at (6.8,-4.525) {$P_\phi(\hat{x})$};
  \end{tikzpicture}
  
   \vspace{-1em}
   \caption{\textbf{D2A-HMR model architecture}. Given an image ($\textbf{I}$), we first incorporate a transformer backbone ($\textbf{\textit{E}}$) to estimate the depth map ($\textbf{D}$) and a CNN backbone ($\textbf{\textit{F}}$) to extract the features from the images. Positional embedding is applied to both image and pseudo-depth features, utilizing a hybrid approach for image tokens ($z_{img}$) and pseudo-depth tokens ($z_{depth}$). Self-attention is performed on $z_{img}$ and $z_{depth}$, resulting in $z_{img}'$ and $z_{depth}'$, respectively. Subsequently, cross-attention is applied between $z_{img}'$ and $z_{depth}'$ to produce $z_c$. The learnable fusion gates combine $z_{img}'$, $z_{depth}'$, and $z_c$, followed by layer normalization and an MLP. The resulting gated tokens ($z$) are input into three distinct refinement modules: a decoder ($\textbf{\textit{D}}$) for silhouette estimation, a regressor head, $\textbf{\textit{R}}$ which incorporates normalizing flow ($\textbf{DM}$) for distribution-aware joint vertex estimation and masked modeling for enhanced semantic representation of the features.}
   \label{fig:overview}
\end{figure*}

\section{Related Work}\label{sec:related-work}

\textbf{Human Mesh Recovery from a Single Image.} Recent works on HMR can be split into parametric and nonparametric approaches. Parametric approaches can further be split into optimization-based and learning-based approaches. Optimization-based approaches fit a body model by minimizing the error between different prior terms. SMPLify \cite{smplify} fits the parametric SMPL \cite{smpl} model to minimize the error between the recovered mesh and keypoints. In addition, prior terms including silhouettes \cite{imphmr, econ} or distance functions \cite{kbody} are used to penalize unrealistic shapes and poses. Learning-based approaches take advantage of deep neural networks to predict model parameters \cite{hmr, spin, imphmr, hmrvit}. 
Recent works including HMR-ViT \cite{hmrvit} use a transformer-only temporal architecture to predict the model parameters, and ImpHMR \cite{imphmr} uses neural feature fields to model humans in 3D space from a single image. 

For directly regressing the vertices, works including GraphCMR \cite{graphcmr}, Pixel2mesh \cite{pixel2mesh}, and Feastnet \cite{feastnet} use graphical neural networks to regress the vertices of RGB images, effectively by modeling neighborhood vertex-vertex interactions. Pose2Mesh \cite{pose2mesh} uses a 2D and 3D pose to regress the vertices using graphical spectral neural networks. METRO \cite{metro} uses transformers to model the global interaction between the vertices and I2LMeshNet \cite{i2lmeshnet} uses a heatmap-based representation called lixel to regress the human mesh. 

\textbf{Normalizing flow.} Normalizing flow is a tool for efficiently transforming a simple distribution into a complex one through a series of invertible transformations \cite{Kirichenko2020WhyNF, prohmr}. It applies to probability density estimation, which can be used to estimate the likelihood. Previous work including \cite{nf_smpl1} and \cite{nf_smpl2} use normalizing flows to learn a priori the distributions of plausible human poses. ProHMR \cite{prohmr} focuses on modeling the output of the human mesh as a distribution over all the different possible meshes. However, it utilizes normalizing flows to directly predict the exact underlying distribution which is demonstrated to perform poorly for OOD data \cite{Kirichenko2020WhyNF}. RLE \cite{rle} uses normalizing flow to minimize the difference between the distributions of the ground truth and predicted 2D poses rather than using the output distribution to sample one particular pose, thereby boosting the performance of regression-based pose estimation techniques.

Inspired by the literature on residual log-likelihood in 2D human pose estimation \cite{rle} and the shortcomings of existing HMR approaches, our approach focuses on mitigating distribution discrepancies of the output and ground truth meshes by leveraging normalizing flow techniques. This alleviates the problem of poor performance on OOD data as we use normalizing flows in the refinement module to minimize the difference between the output mesh distribution and ground truth mesh distribution instead of predicting output poses/ meshes using the captured output distribution.

\textbf{Attention for Human Mesh Recovery.} Attention mechanisms have been shown to be effective for HMR by enabling models to focus on the most relevant parts of the input data. METRO \cite{metro} uses self-attention to reduce ambiguity by establishing non-local feature exchange between visible and invisible parts with progressive dimensionality reduction. SAHMR \cite{sahmr} uses cross-attention between image and scene contact information to improve the posture of the regressed mesh. The recently proposed JOTR \cite{jotr} uses self-attention to study the dependencies of 2D and 3D features to solve problems of occlusion. PSVT \cite{psvt} uses a spatiotemporal attention mechanism to capture relations between tokens and pose/ shape queries in both temporal and spatial dimensions. Similarly, OSX \cite{osx} uses a component-aware encoder to capture the correlation between different parts of the human body to predict the whole-body human mesh. 

We propose a parallel network composed of two self-attention modules to learn global dependencies within the image and pseudo-depth features, respectively, and a cross-attention module to learn inter-modal dependencies between the image and pseudo-depth features. This allows the network to learn a more comprehensive representation for accurate 3D mesh recovery.

\section{Method} \label{sec:method}

The overview of the proposed D2A-HMR framework is presented in Figure \ref{fig:overview}. In this section, we delve into the architecture and training objective of D2A-HMR. The feature encoding process begins with the extraction of features from the image and pseudo-depth map using a convolutional neural network (CNN) backbone, followed by hybrid position encoding. These encoded features are then inputted into the transformer encoder, which engages in cross-attention with the pseudo-depth cues and the input image. Following this, the refinement module comes into play, incorporating the distribution matching, silhouette decoder, and masked modeling components to regularize the model during the training process.


\subsection{Architecture} \label{sec:arch}

\textbf{Feature Encoding.} The initial step involves passing the input image and depth map through a CNN backbone to extract pertinent features. Subsequently, to explicitly model the structure of the features, position embedding is applied to these extracted features. 

Specifically, we implement a hybrid positional encoding ($P_e$) illustrated in Equation \eqref{eq:pe}, for the image and depth tokens. This hybrid approach capitalizes on the strengths of both learnable position embeddings ($P_{l}$) and sinusoidal position embeddings ($P_{s}$). $P_{l}$ adapts to task-specific positional patterns, proving highly effective in capturing intricate spatial relationships. Meanwhile, $P_{s}$ contributes to the globally consistent positional understanding, capturing more information about the position. This combination optimally balances adaptability and global context, yielding fine-grained spatial patterns and general positional relationships.

\begin{equation}
    P_e = \omega_1 P_{l} + \omega_2 P_{s}
    \label{eq:pe}
\end{equation}

where $\omega_1$ and $\omega_2$ are learnable parameters controlling the position embedding contribution of both types. 

\textbf{Transformer Encoder.} The utilization of the transformer encoder in D2A-HMR is driven by the overarching goal of effectively learning pseudo-depth cues from the input data. Using self-attention mechanisms on the encoded features derived from both modalities (image and pseudo-depth map), namely $z_{img}$ and $z_{depth}$, the transformer encoder facilitates understanding of spatial relationships within each domain. Furthermore, we propose to use a cross-attention mechanism to establish intricate connections between the image and pseudo-depth information. The resulting fused representation, denoted as $z$, encapsulates rich depth cues, crucial for the subsequent regression of human vertices.

The embedded features, denoted as $z_{img}$ and $z_{depth}$, serve as input tokens to the transformer encoder, embodying our pursuit of learning pseudo-depth cues. Using self-attention mechanisms, the encoder refines $z_{img}$ and $z_{depth}$ by capturing spatial relationships within each modality, producing updated features $z_{img}'$ and $z_{depth}'$, respectively. Subsequently, the introduction of a cross-attention mechanism facilitates connections between image and pseudo-depth features. The resulting cross-attended tokens denoted as $z_c$, are then fused with $z_{img}'$ and $z_{depth}'$ from their respective attention heads, yielding a final fused representation denoted as $z$, as illustrated in Equation \eqref{eq:output_z}. To facilitate this fusion, learnable fusion gates are employed, similar to the position encoding methodology. These gates adaptively emphasize the importance of each source, enhancing the model's capacity to capture meaningful relationships between the image and pseudo-depth features. 

\begin{equation}
    z = \omega_3 z_{img}' + \omega_4 z_{depth}' + (1-\omega_3-\omega_4)z_{c}
    \label{eq:output_z}
\end{equation}

Here, in Equation \ref{eq:output_z}, $\omega_3$ and $\omega_4$ are the learnable parameters. Once the fusion is done, $z$ is normalized and fed as input to an MLP to get the output tokens. This holistic approach enables our model to effectively capture intricate patterns and dependencies within the input image and the 3D information of the scene. A visual illustration of the transformer encoder is shown in Figure \ref{fig:overview}.

\subsection{Refinement Module} \label{sec:refinement}

The refinement module in the D2A-HMR framework encompasses three key components, each designed to enhance the model's capabilities in capturing different aspects of human pose and shape. First, the distribution matching component aids in refining the model's representation by aligning the output mesh distribution to the ground truth mesh distribution. This adaptation enables the model to capture and adapt to inherent variations in the distribution of training data, promoting a more generalized performance that extends beyond the specific characteristics of the training data. The second component, the silhouette decoder, focuses on optimizing the model's capacity to align the shape with the input image by adeptly capturing the outlines of the human subject. This component contributes significantly to the model's ability to refine and improve its representation based on the visual cues present in the input data. Lastly, the masked modeling component serves to empower the model by learning from available information, thereby enhancing its ability to capture long-range relationships among features in the image. This integration ensures that the model can leverage relationships across the entire input, contributing to a more comprehensive understanding of the underlying human pose and shape.

\textbf{Distribution Matching.} To align the model with the underlying data distribution, we incorporate the RealNVP \cite{realnvp} normalizing flow mechanism within the D2A-HMR framework. This aims to refine the model by minimizing the discrepancy between predicted and ground truth mesh distributions. The transformer encoder's output tokens z are passed through a MLP regressor ($\textbf{\textit{R}}$), which utilize linear layers to predict the mean $\mu$ and standard deviation $\sigma$, controlling the position and scale of the initially assumed Gaussian distribution. The flow-modeled distribution ($P_\phi(\hat{x})$, where $\hat{x}$  is the predicted mesh) is deconstructed into three essential terms, as expressed in the equation:

\begin{equation}
    \log P_\phi(\hat{x}) = \log Q(\hat{x}) + \log \dfrac{P(\hat{x})}{c \cdot Q(\hat{x})} + \log c
\end{equation}

The first term, $\log Q(\hat{x})$, quantifies the logarithmic probability of the data under the simple distribution. The second term, $\log \dfrac{P(\hat{x})}{c \cdot Q(\hat{x})}$, represents the residual log-likelihood, serving as the distinction between the log-probability of the data under the optimal underlying distribution and the log-probability under the tractable initial density function. The third term, $\log c$, functions as a normalization constant.

\textbf{Silhouette Decoder.} To optimize shape alignment, we used a specialized decoder to reconstruct silhouettes. Leveraging features from the transformer encoder, this decoder employs a sequence of deconvolution layers with ReLU activation and dropout, culminating in a fully connected layer. This reconstruction process significantly augments the model's capability to generate high-quality silhouette representations. To acquire the pseudo-ground truth silhouette of human subjects, we utilize an existing segmentation technique \cite{rvm}.

\textbf{Masked Modeling.} Prior works including \cite{bert}, \cite{metro}, and \cite{mim}  have demonstrated the efficacy of masked modeling in elucidating diverse relationships within training datasets, spanning textual, vertex, and image domains respectively. In alignment with these established works, we adopt random masking of the embedded features to recover the vertex of the human body. By deliberately obscuring a percentage of embedded features during training, our model is forced to rely solely on the unmasked features extracted from the image. This enables a comprehensive understanding of both short and long-range relationships among the features, contributing to the overall performance of D2A-HMR framework.

\subsection{Loss Functions} \label{sec:loss}

In this sub-section, we present the comprehensive training objectives employed to recover the human mesh in our model. These objectives consist of a weighted combination of various loss components, each serving a specific role in refining the model's output.

The loss function $\mathcal{L}_v$ is computed using the loss metric $L_1$, with the aim of minimizing the disparities between the model's output vertices with the ground truth vertice representation. Simultaneously, ${\mathcal{L}}_{3D} = | J_{3D} - J^g_{3D} |$ leverages the same loss metric to optimize the 3D pose by regression ($J_{3D}$) of the output mesh vertices following \cite{metro}, seeking alignment with the ground truth pose coordinates ($J^g_{3D}$). To enhance the alignment between image and mesh representations, camera parameters are employed to reproject and infer the 2D human pose coordinates ($J_{2D}$) represented with $\mathcal{L}_{2D} = | J_{2D} - J^g_{2D} |$, where $J^g_{2D}$ is the 2D pose ground truth. This reprojected output is refined by applying loss optimization using $L_1$.


As mentioned in Section \ref{sec:refinement}, a distribution matching regularizer is used to penalize the model for predicting outputs that are unlikely under the underlying ground truth distribution. Equation \eqref{eq:dist_loss} shows the distribution regularizer ($\mathcal{L}_{RLE}$) used in the D2A-HMR architecture. 

\begin{equation}
    \mathcal{L}_{RLE} = - \log Q(\bar{\mu}_g) - \log G_\phi(\bar{\mu}_g) - \log c + log \ \sigma
    \label{eq:dist_loss}
\end{equation}

Here, in Equation \eqref{eq:dist_loss}, $G_\phi(\bar{\mu_g})$ is the learned residual distribution of the predicted value $\bar{\mu_g}$ where $\bar{\mu_g}$ = $(\mu_g - \mu)/ \sigma$. Here, $\mu_g$ is the ground truth distribution and $\phi$ is the flow model parameter. Additionally, we incorporate silhouette loss, denoted $\mathcal{L}_{silh}$, which regularizes the model by controlling the shape of the reconstructed mesh. The overall objective function is shown in Equation \eqref{loss}, which represents a combination of these individual losses. 

\begin{equation}
    \mathcal{L} = \lambda_{d} \mathcal{L}_{RLE} + \lambda_v \mathcal{L}_v + \lambda_{3D} \mathcal{L}_{3D} + \lambda_{2D} \mathcal{L}_{2D} + \lambda_{s} \mathcal{L}_{silh}
    \label{loss}
\end{equation}

where $\lambda_{d}$, $\lambda_v$, $\lambda_{3D}$, $\lambda_{2D}$ and $\lambda_{s}$ denote the weights attributed to the training objectives concerning the distribution, vertices, 3D pose coordinates, 2D pose coordinates, and silhouettes, respectively. 

\section{Experiments}\label{sec:exp}

\subsection{Implementation Details}

\textbf{Training Details.} Training was carried out on an infrastructure comprising three NVIDIA A6000 GPUs. The network was trained for 500 epochs, with a batch size of 48, and 24 parallel workers. Adam optimizer, configured with a learning rate of $10^{-4}$ and beta values of $0.9$ and $0.99$, was used for optimization. The network was designed to output a coarse mesh representation containing $431$ vertices. This output was subsequently upsampled \cite{graphcmr} to the original mesh's $6890$ vertices, utilizing learnable MLP layers, resulting in the model's ability to capture fine-grained spatial details.

\begin{table*}
  \centering
  \caption{Comparison to state-of-the-art 3D pose reconstruction approaches on 3DPW and Human3.6M datasets. \textbf{Bold}: best; \underline{Underline}: second best}
  \begin{tabular}{@{}clcc|cccc@{}}
    \toprule
    & \multirow{2}{*}{\textbf{Method}} & \multicolumn{2}{c|}{\textbf{Human3.6M}} & \multicolumn{3}{c}{\textbf{3DPW}} \\
    \cmidrule(lr){3-4} \cmidrule(lr){5-7}
    & & \textbf{mPJPE} $\downarrow$ & \textbf{PA-mPJPE} $\downarrow$ & \textbf{mPVE} $\downarrow$ & \textbf{mPJPE} $\downarrow$ & \textbf{PA-mPJPE} $\downarrow$\\
    \midrule
    \multirow{3}{*}{\rotatebox{90}{\textbf{Video}}}
    & HMMR \cite{hmmr} & - & 58.1 & 139.3 & 116.5 & 72.6 \\
    & TCMR \cite{tcmr} & 62.3 & 41.1 & 111.5 & 95.0 & 55.8 \\
    & VIBE \cite{vibe} & 65.6 & 41.4 &99.1 & 93.5 & 56.5 \\
    \midrule

    \multirow{7}{*}{\rotatebox{90}{\textbf{Model-based}}}
    & HMR \cite{hmr} & 88.0 & 56.8 & - & 130.0 & 81.3   \\
    & SPEC \cite{spec} & - & - & 118.5 & 96.5 & 53.2  \\
    & SPIN \cite{spin} & 62.5 & 41.1 & 116.4 & 96.9 & 59.2  \\
    & PyMAF \cite{pymaf} & 57.7 & 40.5 & 110.1 & 92.8 & 58.9  \\  
    & ROMP \cite{romp} & - & - & 105.6 & 89.3 & 53.5\\
    & HMR-EFT \cite{hmreft} & 63.2 & 43.8 & 98.7 & 85.1 & 52.2  \\
    & PARE \cite{pare} & 76.8 & 50.6 & 97.9 & 82.0 & 50.9  \\
    \midrule

    \multirow{6}{*}{\rotatebox{90}{\textbf{Model-free}}}
    & ProHMR \cite{prohmr} & - & 41.2 & 109.6 & 95.1 & 59.5  \\
    & I2LMeshNet \cite{i2lmeshnet} &  55.7 & 41.1 & - & 93.2 &  57.7   \\
    & Pose2Mesh \cite{pose2mesh} & 64.9 & 47.0 & - & 89.2 & 58.9  \\
    & METRO \cite{metro} & \underline{{54.0}} & \underline{{36.7}} & \textbf{{88.2}} & \textbf{77.1} & {\textbf{47.9}}  \\
    \cmidrule(lr){2-7} 
    
    & \textbf{D2A-HMR (Ours)} & \textbf{53.8} & \textbf{36.2} & \underline{{88.4}} & \underline{{80.5}} & \underline{{48.4}} \\

    \bottomrule
  \end{tabular}
  \label{tab:sota}
\end{table*}

\textbf{Datasets.} Following previous work, we used two prominent 3D human pose estimation datasets, namely 3D Poses in the Wild (3DPW) \cite{3dpw} and Human3.6M \cite{h36m} to train our D2A-HMR model. For the 3DPW dataset, we follow the standard practice of splitting the dataset into a training set of 22,000 images and a test set of 35,000 images. In the case of Human3.6M, we trained our D2A-HMR model on subjects S1, S5, S6, S7, and S8 and conducted testing on subjects S9 and S11. These data configurations were aligned with the common training and evaluation settings within the domain \cite{metro, hmmr}. The qualitative evaluation of the model was done in Leeds Sports Pose (LSP) \cite{lsp}, and various dedicated sports datasets including the MLBPitchDB dataset \cite{mitigatingblur} and HARPE dataset \cite{icehockey}.

\textbf{Evaluation Metrics.} In line with established practices from previous research \cite{pare, metro, pose2mesh}, we subjected our model to a comprehensive evaluation using key metrics: mean per joint position error (mPJPE), procrustes-aligned mean per joint position error (PA-mPJPE) and per vertex error (mPVE) in both the 3DPW and Human3.6M datasets. mPVE metric is ignored if the ground truth mesh is not available. All metrics were measured in millimeters (mm), providing a precise assessment of our model's performance.

\subsection{Main Results}

We assess the performance of the proposed D2A-HMR framework by comparing it with established state-of-the-art techniques for HMR. The results, presented in Table \ref{tab:sota}, highlight the competitive performance of our method across various metrics on the Human3.6M and 3DPW datasets. The comparative results demonstrate that the meshes generated by the D2A-HMR framework exhibit superior alignment with the input image. Our method's adept understanding of pseudo-depth cues and the distribution contributes significantly to improved alignment, particularly in handling challenging input scenarios characterized by depth ambiguities and extreme poses. 



\begin{table}[H]
  \centering
  \caption{Comparison of D2A-HMR on a baseball dataset \cite{mitigatingblur}. \textbf{Bold}: best; \underline{Underline}: second best; \underline{\underline{Double Underline}}: third best.}
  \begin{tabular}{@{}lccc@{}}
    \toprule
    \textbf{Method} & \textbf{Acc.} $\uparrow$ & \textbf{mPJPE} $\downarrow$ \\
    \midrule
    HMR \cite{metro} & 65.9 & 61.3 \\
    SPIN \cite{spin} & \underline{84.7} & \underline{32.1} \\
    ProHMR \cite{metro} & 76.1 & 48.2 \\
    ROMP \cite{metro} & 77.4 & 48.9 \\
    METRO \cite{metro} & 81.5 & 37.8 \\
    PARE \cite{pare} & \underline{\underline{84.0}} & \underline{\underline{33.7}} \\
    \midrule
    \textbf{D2A-HMR (Ours)} & \textbf{87.9} & \textbf{30.6}  \\
    \bottomrule
  \end{tabular}
  \label{tab:itw}
\end{table}

Table \ref{tab:itw} presents a comprehensive comparison between our proposed method and established state-of-the-art HMR techniques, utilizing the baseball dataset \cite{mitigatingblur}. Notably, D2A-HMR demonstrates \textit{superior performance} in terms of accuracy and mPJPE on this dataset, which is characterized by \textit{high player motion blur and instances of self-occlusion}. 

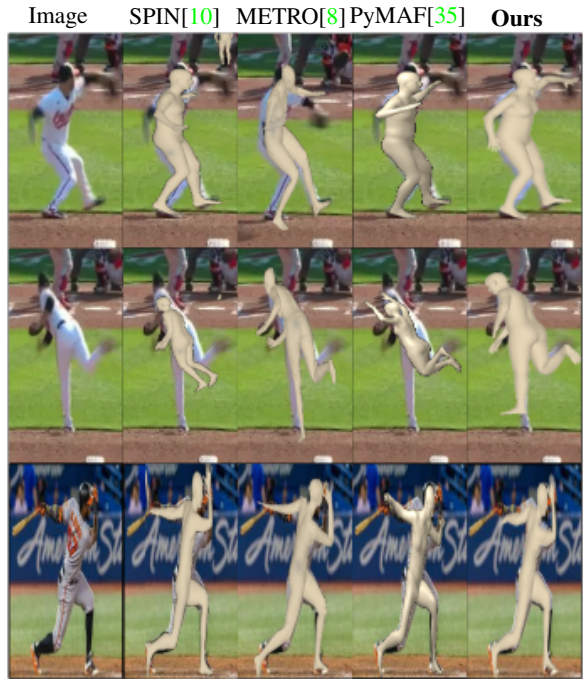
\begin{figure}[H]
  \centering
  \begin{tikzpicture}
    \node at (0,0) {\includegraphics[width=1\linewidth]{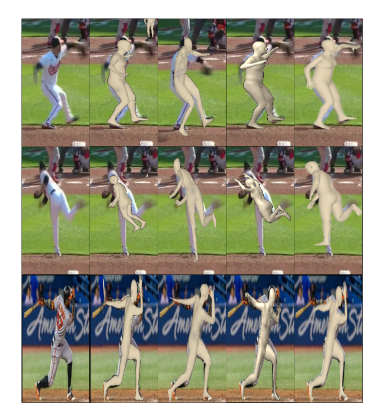}};
    \node[font=\small] at (-3.15,4.3) {$\text{Image}$};
    \node[font=\small] at (-1.6,4.3) {$\text{SPIN\cite{spin}}$};
    \node[font=\small] at (-0.05,4.3) {$\text{METRO\cite{metro}}$};
    \node[font=\small] at (1.5,4.3) {$\text{PyMAF\cite{pymaf}}$};
    \node[font=\small] at (2.95,4.3) {$\textbf{Ours}$};
  \end{tikzpicture}
  \vspace{-13px}
  \caption{\textbf{Qualitative results.} Inferred SMPL mesh reconstruction on the MLBPitchDB baseball dataset \cite{mitigatingblur}.}
  \label{fig:qualitative-baseball}
\end{figure}

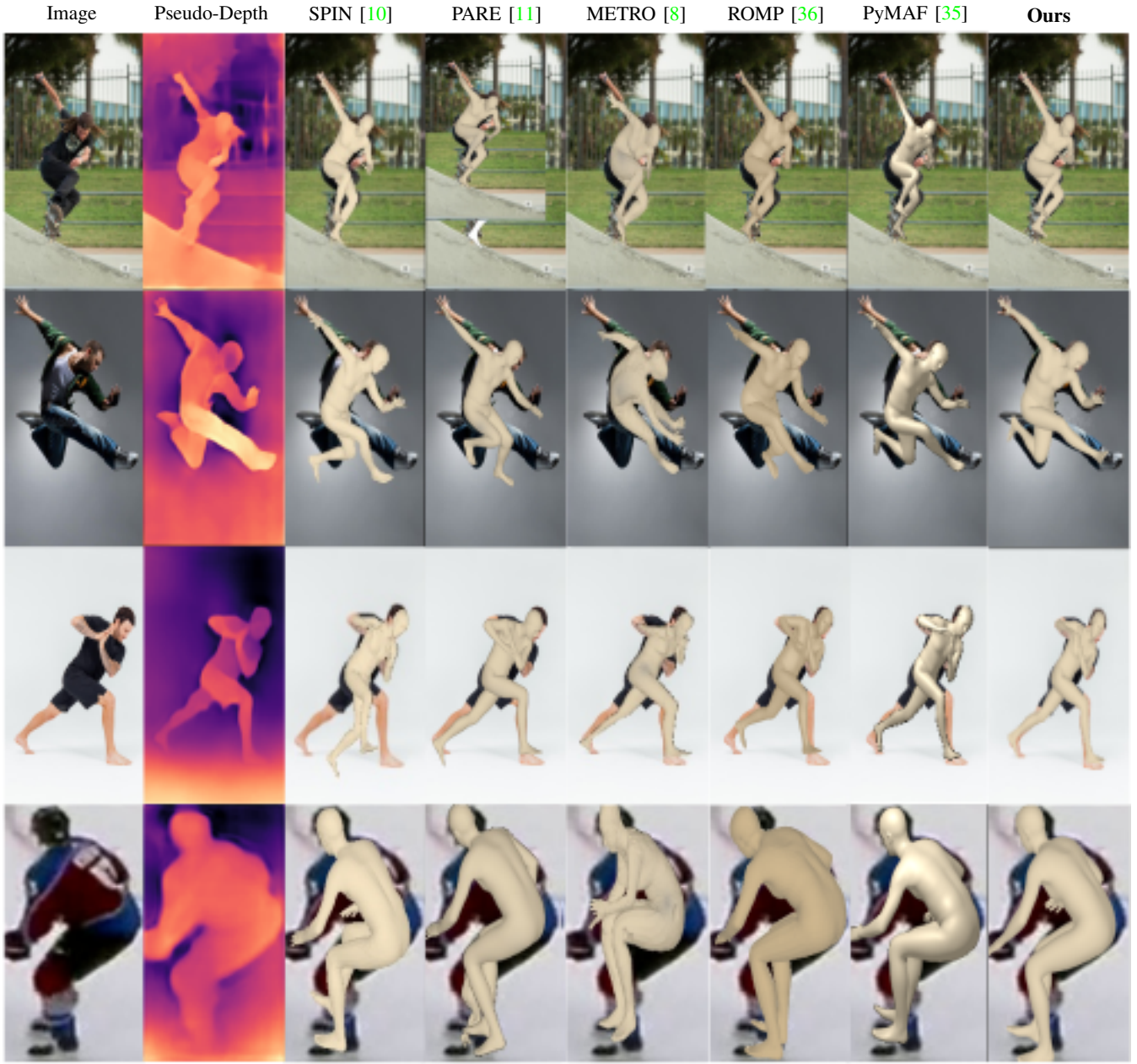
\begin{figure*}
  \centering
  \begin{tikzpicture}
    \node at (0,0) {\includegraphics[width=\linewidth]{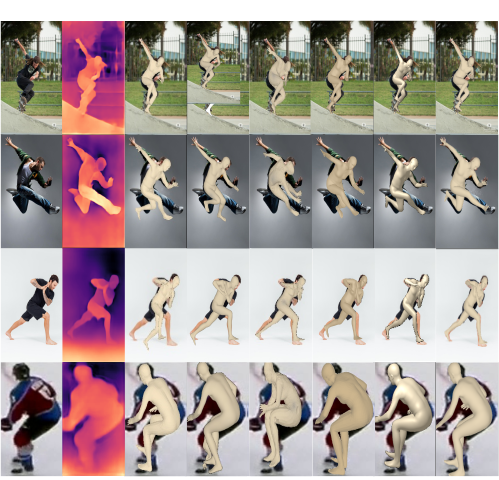}};
    \node[font=\small] at (-7.8,8.1) {$\text{Image}$};
    \node[font=\small] at (-5.6,8.1) {$\text{Pseudo-Depth}$};
    \node[font=\small] at (-3.4,8.1) {$\text{SPIN \cite{spin}}$};
    \node[font=\small] at (-1.1,8.1) {$\text{PARE \cite{pare}}$};
    \node[font=\small] at (1.1,8.1) {$\text{METRO \cite{metro}}$};
    \node[font=\small] at (3.3,8.1) {$\text{ROMP \cite{romp}}$};
    \node[font=\small] at (5.5,8.1) {$\text{PyMAF \cite{pymaf}}$};
    \node[font=\small] at (7.6,8.1) {$\textbf{Ours}$};
  \end{tikzpicture}
  \vspace{-18px}
  \caption{\textbf{Qualitative results.} Qualitative comparison of D2A-HMR with SPIN \cite{spin}, PARE \cite{pare}, METRO \cite{metro}, ROMP \cite{romp} and PyMAF \cite{pymaf} on in-the-wild data from different sports dataset \cite{mitigatingblur, icehockey, lsp} and unusual poses from the internet.}
  \label{fig:qualitative}
\end{figure*}

Figure \ref{fig:qualitative-baseball} visualizes the effectiveness of our approach in handling these complexities, highlighting its robustness to unseen poses. To further emphasize the efficacy of our proposed approach, we conducted a qualitative comparison against several state-of-the-art techniques on unseen poses, as depicted in Figure \ref{fig:qualitative}. This comparative analysis underscores D2A-HMR's potential for tackling challenging real-world scenarios.

\subsection{Ablation Studies}

To verify the individual impact of each module on the proposed D2A-HMR model, comprehensive studies were conducted, as detailed in this sub-section. For consistency across all studies, the 3DPW dataset was utilized as the common benchmark.

\textbf{Integration of multi-modal data.} Experimentation to assess the impact of depth and distribution matching components within the D2A-HMR are detailed in Table \ref{tab:modules}.

\begin{table}[H]
  \centering
  \caption{Ablation study on pseudo-depth and distribution modeling for D2A-HMR evaluated on 3DPW dataset}
  \begin{tabular}{@{}cccc@{}}
    \toprule
    \textbf{Depth} & \textbf{Dist.} & \textbf{mPJPE} $\downarrow$ & \textbf{PA-mPJPE} $\downarrow$ \\
    \midrule
    \ding{51} & & 92.7 & 61.8 \\
    & \ding{51} & 90.0 & 56.9 \\
    \ding{51} & \ding{51} & \textbf{80.5} & \textbf{48.4} \\
    \bottomrule
  \end{tabular}
  \label{tab:modules}
\end{table}

Incorporation of both the pseudo-depth and distribution modeling modules in the D2A-HMR framework is observed to lead to a substantial improvement in the overall performance of mesh recovery. This observation serves as confirmation that the underlying motivation behind the proposed framework is valid and aids in enhancing the model's capabilities.

\textbf{Depth on mPJPE(\textit{z}).} Experimentation on exclusively capturing the depth component of the regressed 3D joints in order to demonstrate its impact on the human pose is conducted in Table \ref{tab:ablation-depth}.

\begin{table}[H]
  \centering
  \caption{Ablation study on the impact of depth modeling for D2A-HMR evaluated on 3DPW dataset}
  \begin{tabular}{@{}lcc@{}}
    \toprule
     & \textbf{mPJPE(\textit{z})} $\downarrow$ & \textbf{PA-mPJPE(\textit{z})} $\downarrow$ \\
    \midrule
    w/o depth modeling & 69.1 & 58.3 \\
    w/ depth modeling & \textbf{65.4} & \textbf{53.6} \\
    \bottomrule
  \end{tabular}
  \label{tab:ablation-depth}
\end{table}

A notable enhancement in the z-axis of the reconstructed mesh is evident, as highlighted in Table \ref{tab:ablation-depth}. We computed mPJPE along the z-axis denoted as mPJPE($\textit{z}$), disregarding the components $x$ and $y$ of the reconstructed mesh. This experimentation validates that the incorporation of scene-depth information contributes to an improvement in HMR.

\textbf{Silhouette and Masked Modeling.} Table \ref{tab:modules-ablation} illustrates the impact of the silhouette decoder and masked modeling used within the D2A-HMR framework. 

\begin{table}[H]
  \centering
  \caption{Ablation study on the silhouette decoder and masked modeling evaluated on 3DPW dataset}
  
  \begin{tabular}{@{}cccc@{}}
    \toprule
    \textbf{Silhouette} & \textbf{Masked Modeling} & \textbf{mPJPE} $\downarrow$ & \textbf{PA-mPJPE} $\downarrow$ \\
    \midrule
    \ding{51} & & 89.5 & 62.2 \\
    & \ding{51} & 84.7 & 51.4 \\
    \ding{51} & \ding{51} & \textbf{80.5} & \textbf{48.4} \\
    \bottomrule
  \end{tabular}
  \label{tab:modules-ablation}
\end{table}

The observations drawn from Table \ref{tab:modules-ablation} highlight the beneficial impact of incorporating both the silhouette decoder and masked modeling modules in enhancing the model's ability to disentangle the appearance and part-relationship of the person. While prior studies, such as \cite{pymaf}, employ methods like explicit iterative optimization for mesh-to-image alignment, our silhouette decoder yields improved alignment outcomes compared to scenarios without the decoder. Thus, these modules are utilized during the training process of the D2A-HMR framework, contributing to its improved performance.

\textbf{Backbones.} A comprehensive analysis of D2A-HMR's performance by investigating its behavior with various backbone architectures was conducted. To establish a strong baseline, we first trained two ResNet variants for 1000 epochs on the ImageNet dataset \cite{imagenet} for an image classification task. We also explored HRNet variants trained for 1000 epochs using the COCO dataset \cite{coco} for the classification task.

\begin{table}[H]
  \centering
  \caption{Different input representations as the backbone for D2A-HMR evaluated on 3DPW dataset}
  \begin{tabular}{@{}lcc@{}}
    \toprule
    \textbf{Backbone} & \textbf{mPJPE} $\downarrow$ & \textbf{PA-mPJPE} $\downarrow$ \\
    \midrule
    ResNet50 & 91.1 & 59.9 \\
    ResNet101 & 89.5 & 55.8 \\
    HRNet-w40 & 85.2 & 52.1 \\
    HRNet-w64 & \textbf{80.5} & \textbf{48.4} \\
    \bottomrule
  \end{tabular}
  \label{tab:input}
\end{table}

We observe that HRNet-w64 gives the most positive impact on feature extraction from both the image and depth maps compared to the ResNet backbones. This can be attributed to HRNet-w64's effectiveness in capturing both local and global contexts through its multiresolution fusion representations, thereby enhancing the model's ability to extract rich and informative features.

\section{Conclusion}\label{sec:discussion}

In summary, our research introduces the Distribution and Depth-Aware Human Mesh Recovery (D2A-HMR) framework as an innovative solution to the persistent challenge of depth ambiguities and distribution disparities in monocular human mesh recovery. By explicitly incorporating scene-depth information, we have substantially reduced the inherent ambiguity, resulting in a more precise and accurate alignment of human meshes. The utilization of normalizing flows to model the output distribution has been instrumental in regularizing the model to minimize the underlying distribution disparities, enhancing its resilience against noisy labels, and mitigating biases in human-form modeling. 

Our extensive experimentation on diverse datasets has demonstrated the competitive performance of the D2A-HMR method when compared to state-of-the-art HMR techniques. Furthermore, it has been noticed that our network outperforms existing work on sports datasets with OOD data. This proposed framework not only addresses depth ambiguities and mitigates noise, but also leverages the inherent 3D information present in images, providing a robust and unambiguous solution for human mesh recovery. Future work will entail training on more diverse datasets to enhance the alignment and generalizability of the HMR process. 


\section*{Acknowledgement}

We extend our gratitude to the Baltimore Orioles of Major League Baseball, whose generous support through the Mitacs Accelerate Program played a pivotal role in advancing this research. We also acknowledge the Digital Research Alliance of Canada for their invaluable hardware support.

\bibliographystyle{IEEEtran}
\bibliography{main.bib}
\end{document}